# Integrating Logical and Probabilistic Reasoning for Decision Making

Jack Breese
Rockwell International Science Center
Palo Alto Laboratory
444 High Street, Suite 400
Palo Alto, California 94301

Edison Tse
Department of Engineering-Economic Systems
Stanford University
Stanford, California 94305

## ABSTRACT

We describe a representation and a set of inference methods that combine logic programming techniques with probabilistic network representations for uncertainty (influence diagrams). The techniques emphasize the dynamic construction and solution of probabilistic and decision-theoretic models for complex and uncertain domains. Given a query, a logical proof is produced if possible; if not, an influence diagram based on the query and the knowledge of the decision domain is produced and subsequently solved. A uniform declarative, first-order, knowledge representation is combined with a set of integrated inference procedures for logical, probabilistic, and decision-theoretic reasoning.

## I. INTRODUCTION

Recent advances in representation and inference under uncertainty for artificial intelligence have stressed the utility of network representations. Alternative representations for probabilistic inference include influence diagrams, developed by decision analysts (Howard and Matheson, 1981), and the related formalism of Bayes networks (Pearl, 1986). Though there appears to be some agreement on graphical depictions of dependencies for uncertainty in AI, much less attention has been devoted to the generation or construction of these structures. Most researchers focus on the procedures for propagating information and manipulating structures for given diagram (Pearl, 1986; Henrion, 1986; Shachter, 1986b; Shenoy, Shafer, and Melloui, 1986).

This paper develops techniques designed to allow reasoning about the structure of a probabilistic or decision-theoretic model as opposed to reasoning with a given model. The objective is to provide a representation, a set of inference techniques, and an architecture which can support dynamic construction and solution to a probabilistic model in response to a query and domain knowledge. Given a query, the basic idea is to produce a logical proof if possible; if not, the knowledge of the decision domain is searched to find information which defines a probabilistic or decision-theoretic model for the query. This model is produced and subsequently solved. The approach developed here has the following advantages over previous approaches:

- Probabilistic reasoning is gracefully integrated with logical, deterministic inference. This allows one to invoke the appropriate richness of representation for different problems based on information availability and desired solution methodology.

- The expressiveness of the language does not impose assumptions of conditional independence on the probabilistic representation. The knowledge base can therefore express the set of dependencies and independencies made explicit by the system builder and/or decision maker.

- Through the dynamic construction of models in response to queries and as the state of information in the knowledge base changes, the size of probabilistic models is minimized.

- The system is capable of construction of multiple models for the same phenomena. This allows reasoning about the performance and results of different models within the same environment.

- By formulating logical, probabilistic, and decision-theoretic inference within an integrating framework, techniques of explanation and heuristic search can be applied to the construction of probabilistic and decision-theoretic models.

## II. REPRESENTING DECISION DOMAINS WITH PROBABILITIES

In this section a declarative language based on first-order predicate calculus is described for representing decision domains. The language allows the expression of the logical and probabilistic relationships in the domain, as well as the information flows, alternatives, and objectives inherent in decision making contexts. Influence diagrams, graphical depictions for decision problems used in decision analysis, are then briefly presented.

### A. Propositions

The decision domain is represented with a set of propositions of the form (P $x_1$ $x_2$ ..$x_n$) where the P is a relational constant and the $x_i$ are variables or object constants. Given the overall structure of the proposition (i.e. its relational constant and arity), there are three levels of knowledge possible regarding a proposition.

First, there may be a fact stored in the knowledge base regarding regarding a proposition. Thus (WEATHER RAINY SATURDAY) represents the belief under certainty (not subject to updating) that the weather was rainy on Saturday.

At a second level, the possible instantiations of a proposition are restricted to a specified set. This is acheived by associating a set of alternative values with particular variables in a proposition. The alternative values are the set of mutually exclusive, collectively exhaustive values for that variable in the proposition. The alternative values set is a means of restricting the possible values a variable can take on in a proposition.

Consider the WEATHER proposition (WEATHER x y) where x and y are variables. Let x be a restricted variable with its outcomes restricted to the set {FAIR, CLOUDY, RAINY}. Let y be a free, unrestricted variable. Then the only possible values for x in weather are FAIR, CLOUDY, or RAINY. Furthermore for any y, exactly one of the assertions (WEATHER FAIR y), (WEATHER CLOUDY y) , and (WEATHER RAINY y) can be true. These expressions will be referred to as the **alternative outcomes** of the proposition. A proposition with restricted variables is a **restricted proposition** and is written (WEATHER {FAIR, CLOUDY, RAINY} y). The alternative outcomes of a conjunction are the members of the cross product of the alternative outcomes for the component propositions.

The third level of information regarding a restricted proposition is expressed as a measure of belief over the alternative outcomes of the proposition. Specifically, a **probability distribution** maps each alternative outcome to a probability. Since the alternative outcomes are collectively exhaustive and mutually exclusive, the sum of the probabilities is one. The probability distribution for the proposition (WEATHER {FAIR, CLOUDY, RAINY}

RAINY} MONDAY) might be:

| | | |
|---|---|---|
| (WEATHER FAIR MONDAY) | => | 0.7 |
| (WEATHER CLOUDY MONDAY) | => | 0.2 |
| (WEATHER RAINY MONDAY) | => | 0.1 |

Note that probability distributions can only be specified for those propositions for which alternatives sets have been specified. Although a probability provides less information than the assertion of a fact, it requires a larger amount of information in the system since all possible outcomes must be enumerated.

## B. Influences

Influences are well formed formulas in the language and take on one of the following forms: $A \leftarrow B$, $A|_p B \equiv \pi_p(\omega_A | \omega_B)$, or $A|_i B$. All variables which are not restricted are universally quantified. For each formula, A is a single proposition, and B is a conjunction of zero or more propositions.

An influence of the form $A \leftarrow B$ is a **logic influence**. It expresses a standard Horn clause conditional - If B is true then conclude A to be true.

An influence of the form $A|_p B \equiv \pi_p(\omega_A | \omega_B)$ is a **probabilistic influence**. It is a generalization of the logic influence. It expresses the conditional probability distribution for the possible outcomes of the restricted proposition A. Thus, for each alternative outcome of B (written as $\omega_B$), it provides the probability distribution over the alternative outcomes for A (written as $\omega_A$). The first expression $A|_p B$ expresses the existence of a dependence; $\pi_p(\omega_A | \omega_B)$ is the conditional distribution which provides the numerical values for the distribution. Note that in general, B is conjunction of the form $B_1 \wedge B_2 \wedge \dots \wedge B_n$. Some of the $B_i$ are restricted propositions others are not. $\pi_p(\omega_A | \omega_B)$ provides the conditional probability over the alternative outcomes of A given each possible combination of alternative outcomes for the $B_i$ which are restricted propositions. For example,

(WEATHER x TOMORROW) $|_p$ (WEATHER y TODAY)

$\equiv \pi_p(\omega_{\text{(WEATHER x TOMORROW)}} | \omega_{\text{(WEATHER y TODAY)}})$

asserts that the distribution of outcomes for $x \in$ {FAIR,CLOUDY,SUNNY} is provided for each possible outcome of (WEATHER y YESTERDAY). We could condition on more information, perhaps a forecast is also available.

(WEATHER x TOMORROW) $|_p$ (WEATHER y TODAY)$\wedge$(FORECAST z TODAY)

$\equiv \pi_p(\omega_{\text{(WEATHER x TODAY)}} | \omega_{\text{(WEATHER y TODAY)}}, \omega_{\text{(FORECAST z TODAY)}})$

Suppose there is another set of conditions which would change the distribution for (WEATHER x TODAY). If there is a temperature inversion, then an alternative influence can be expressed:

(WEATHER x TOMORROW) $|_p$ (INVERSION TODAY)$\wedge$(WEATHER y TODAY)

$\equiv \pi_p'(\omega_{\text{(WEATHER x TOMORROW)}} | \omega_{\text{(WEATHER y TODAY)}})$

Here (INVERSION TODAY) is treated deterministically. It expresses the condition, assumed to be known with certainty (or not) under which $\pi_p'$ should be used to express the uncertainty

357

regarding (WEATHER x TODAY). If (INVERSION TODAY) is false, then one of the other influences can be used.

An influence of the form A|$_i$ B is an **informational influence**. An influence of this form asserts that the restricted proposition A is a **decision** proposition, i.e. the restricted variables in the proposition are under the control of the decision maker. The alternative outcomes for the proposition are interpreted in this context as the alternative choices facing the decision maker. This influence expresses the informational availability at the time of the decision - specifically the outcomes of the restricted variables in B are known at the time the decision regarding A is made.

(ACTIVITY x TOMORROW) |$_i$ (WEATHER y TOMORROW)

If the alternative values for x are {PICNIC, WORK, SLEEP} then the influence above says that when the decision regarding ACTIVITY is made, we know the outcome for WEATHER.

A decision domain represented in terms of these constructs consists of a set of propositions and influences expressing beliefs, uncertainties, dependencies and information flows.

## C. Influence Diagrams

Influence diagrams are network depictions of decision situations (Howard and Matheson, 1981) developed as a tool for model construction and representation in decision analysis. Each node in the diagram represents an uncertain variable or a decision variable. There is a single node which designated as the value node. This node's variable (real valued) will be maximized or minimized in expected value when solving for optimal decisions. Links between nodes provide a graphical depiction of probabilistic independence and information flows. The formalism of Bayes networks (Pearl, 1986) are similar constructs which express probabilistic dependencies, but do not have a representation for preferences or decisions.

Recently there has been attention devoted to influence diagrams based on providing a complete description of a decision problem. In addition to representing the structure of a decision model, information characterizing the nature and content of nodes and links is attached to the diagram. The diagram then provides a precise specification of a decision maker's preferences, probability assessments, decision alternatives, and states of information. The diagrammatic representations can be directly manipulated to perform probabilistic inference and to generate decision-theoretic recommendations (Shachter,1986a and 1986b). It is this ability to manipulate the diagram directly in order to perform Bayesian inference (equivalent to reversing an arc), forming conditional expectation (equivalent to removing a chance node), and maximize expected utility (equivalent to removing a decision node) which will be utilized in the inference systems developed below.

## III. OVERVIEW OF INFERENCE METHODS

As outlined in the previous section, there are three levels of knowledge regarding a proposition: a fact, a set of possible alternative outcomes for the proposition, and a probability distribution over those alternative outcomes. If assertions in these forms are not explicitly in the knowledge base for a given propositional pattern, we may wish to derive them given the sentences in the knowledge base. Below, this inference process is described for a probabilistic query, i.e. find the probability distribution over alternative outcomes for a proposition (A $t_1$ $t_2$ ....$t_k$) where the $t_i$ are variables or constants. The general scheme is illustrated in Figure 1.

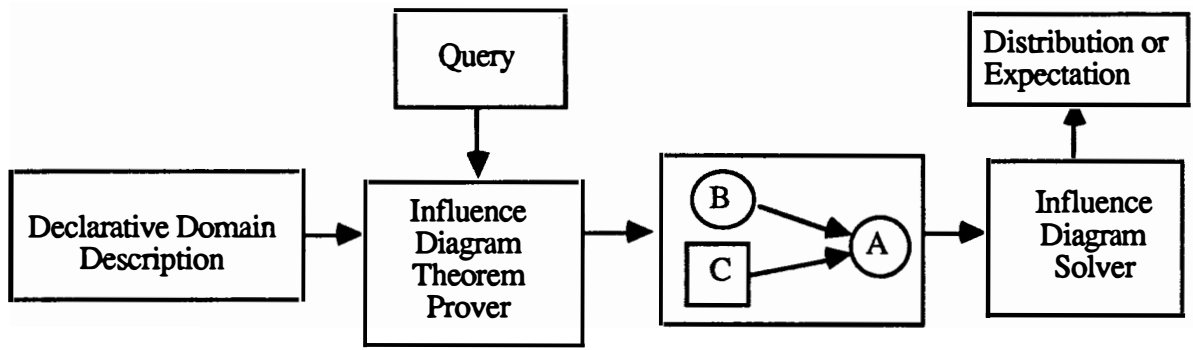

Figure 1 - Overview of Inference

The process starts with a query, for example, "What is the probability distribution over the alternative values for $t_1$ and $t_2$ in (A $t_1$ $t_2$ 1)?". Given this query and the set of sentences comprising the declarative domain description, the inference procedure builds an influence diagram with a node representing the query as its root. This influence diagram is subsequently solved by reducing the diagram to a single node with no predecessors to provide the probability distribution over the alternative outcomes for A, the answer to the query.

## A. Probabilistic Inference

Inference is initiated by a identifying an initial goal (the query), $G_0 = (P\ t_1\ t_2\ ....t_k)$, and an empty influence diagram, N. The proof procedure will search the set of expressions in the decision domain to either 1) logically derive deterministic conclusions regarding the goal or 2) to construct the appropriate probabilistic model that will satisfy the goal. From the initial goal $G_0$, a successful proof will generate a sequence of goals $G_1$ to $G_n$ where $G_n$ is empty. A transformation to a successive goal may add a node to the influence diagram N. The conclusion of a successful proof results in a well-formed influence diagram constructed during the procedure, and an answer substitution $\theta$ providing the bindings on the variables in the original goal.[1] The transformations of goal states proceeds as follows.

Let $G_i$ be a conjunction of the form $P_1 \wedge P_2 \wedge .. \wedge P_k \wedge .... \wedge P_m$, and let N be the set of nodes in an influence diagram. Select some $P_k$ as a subgoal. Then a new goal $G_{i+1}$ can be derived from $G_i$ if one of the following conditions hold.

i.) $P_k$ is logically derivable from the set of formulas of the form A←B by standard Horn clause logical inference, with answer substitution $\theta_{i+1}$. Then

$$G_{i+1} = (P_1 \wedge P_2 \wedge .. \wedge P_{k-1} \wedge P_{k+1} .... \wedge P_m)\theta_{i+1}$$

Thus if a subgoal is known in the database or is provable using a logic proof procedure, it can be removed from the list of subgoals. In this way the procedure described here fully subsumes Horn clause logical inference.

---

[1] A substitution $\theta$ is a set of the form $\{x_1/t_1, x_2/t_2, .....x_n/t_n\}$ where the $x_i$ are variables and the $t_i$ are variables, constants, or alternative value sets as defined in the text. The expression $P\theta$ is the proposition P with $t_i$ substituted for all occurrences of $x_i$ in P. If $P\theta = Q\theta$ then P and Q are said to unify and $\theta$ is a unifier.

ii.) There exists a node $N_j$ in N and a substitution $\theta_{i+1}$ such that $N_j \theta_{i+1} = P_k \theta_{i+1}$. Then

$$G_{i+1} = (P_1 \wedge P_2 \wedge .. \wedge P_{k-1} \wedge P_{k+1} .... \wedge P_m) \theta_{i+1}$$

In this step, we check if the particular subgoal has already been accounted for in the influence diagram N. If there is already a node in the diagram which unifies with the subgoal, then the subgoal can be removed from the list of subgoals. An arc is added to the diagram from $N_j$ to the node created when the subgoal was added to the subgoal list (see Step iv. below).

iii.) There exists a probability distribution $\pi_p(\omega_A)$ for a proposition A and a substitution $\theta_{i+1}$ such that $A \theta_{i+1} = P_k \theta_{i+1}$. Then

$$G_{i+1} = (P_1 \wedge P_2 \wedge .. \wedge P_{k-1} \wedge P_{k+1} .... \wedge P_m) \theta_{i+1}$$

The set of chance nodes in the diagram N is augmented with node labeled $A\theta_{i+1}$ and probability distribution $\pi_p(\omega_A \theta_{i+1})$. The new node represents the prior probability on the proposition, and the node in the diagram has no predecessors. Its successor is that node that was created when the subgoal was added to the subgoal list. A node that encodes the available information about the subgoal has been added to the diagram, therefore the subgoal can be removed from the list of subgoals. This step is analogous to unification with a fact in a logic-based system.

iv.) There exists an influence of the form $A|_p B \equiv \pi_p(\omega_A | \omega_B)$ where B is a conjunction of the form $Q_1 \wedge Q_2 \wedge .. \wedge Q_n$ and a substitution $\theta_{i+1}$ such that $A \theta_{i+1} = P_k \theta_{i+1}$. Then

$$G_{i+1} = (P_1 \wedge P_2 \wedge .. \wedge P_{k-1} \wedge Q_1 \wedge Q_2 \wedge .. \wedge Q_n \wedge P_{k+1} .... \wedge P_m) \theta_{i+1}.$$

This is the backwards chaining step in the procedure. The set of chance nodes in the diagram N is augmented with node labeled $A\theta_{i+1}$ and conditional probability distribution $\pi_p(\omega_A | \omega_B)$, relating A to the restricted variables in B. The new goal, $G_{i+1}$, has been extended with the propositions in the antecedent of the influence. Each of these new subgoals are associated with the new node $A\theta_{i+1}$, since any node created as a result of these new subgoals will have $A\theta_{i+1}$ as a successor.

The probabilistic proof procedure is successful when a derived goal $G_n$ is empty. The answer substitution for the proof is the composition of the substitutions found at each step $\theta_1 \theta_2 ... \theta_n$. The resulting influence diagram N (if non-empty) is then manipulated to derive the desired distribution or expectation. The manipulation consist of removing all nodes in the diagram except that associated with the original goal $G_0$, possibly with some arc reversals (applications of Bayes rule). The probability distribution in the single remaining node is the probability distribution over the alternative outcomes of the original query $(P\ t_1\ t_2\ ....t_k)$.

There are analogous inference techniques for deducing alternative outcomes for propositions and for explicit consideration of decision and value nodes in influence diagrams with this approach (Breese, 1987). For decision making, the initial goal is a proposition with a single real valued variable representing the value to be maximized or minimized. Inference proceeds as above, with the consideration of informational influences $A|_i B$, the addition of decision nodes to the diagram, and a mechanism for associating outcomes of uncertain propositions with a value proposition.

## B. Integrating Logical, Probabilistic, and Decision-theoretic Inference

The representation and inference procedures outlined above integrate logical, probabilistic, and decision-theoretic reasoning in several ways.

First, there is a uniform syntax for logical and probabilistic statements, providing a first-order language for declaratively describing logical and probabilistic relationships as well as information availability for decisions. This allows knowledge bases to freely intermix logical and probabilistic descriptions, without awkward (and frequently incorrect) conversions of one to the other. The representation is modular, in that the addition or deletion single influence does not impose any assumptions concerning conditional independence of propositions as expressed in other influences. This is because a single probabilistic influence in this framework requires consideration of all possible alternative values for both antecedent and consequent. Attachment of a single number to a logic influence, as in the certainty factor representation of uncertainty, does not have this property (Heckerman and Horvitz, 1986). The representation allows the system builder to describe the domain, using both probabilistic (uncertain) and logic (certain) relationships.

Logic also provides not only a means of reasoning about the domain of interest, but also about how to reason about the structure of probabilistic dependencies in the domain. The influence:

(PROPOSITION-OF-INTEREST x) $|_p$ (CONDITIONING-PROPOSITION y)

$$\wedge \text{(CONDITION-FOR-EXPANSION)}$$

$$\equiv \pi(\omega(\text{PROPOSITION-OF-INTEREST x}), \omega(\text{CONDITIONING-PROPOSITION y}))$$

says the probability distribution for PROPOSITION-OF-INTEREST is conditioned on CONDITIONING-PROPOSITION. There is an additional requirement that CONDITION-FOR-EXPANSION be a known fact or provable from the knowledge base. Thus in order to derive the probability distribution of PROPOSITION-OF-INTEREST in terms of CONDITIONING-PROPOSITION, CONDITION-FOR-EXPANSION must be true.[2]

This general scheme allows for reasoning about the structure of probabilistic models in a rule-based manner based on domain or heuristic information. Most other techniques for probabilistic reasoning implicitly rely on a static, prespecified representation of uncertain relationships, e.g., a single Bayes network or influence diagram. Though one can envision a massive global probabilistic model including all propositions, their possible outcomes, and potential dependencies, a model of this type would be extremely cumbersome and inflexible with respect to changes in the model description.

The inference methods admit a precedence in choice of which procedure to use to attempt to address a particular subgoal. This precedence implies a control structure for the search for probabilistic and decision-theoretic models. Overall, control is focused on **minimizing** the extent of models which explicitly account for uncertainty using probability. We first use logic to attempt draw conclusions on any subgoal. If there is information available that asserts the categorical truth value of a proposition, then the other possible values can be ignored. Another level of control involves the selection of alternative probabilistic representations. In the implementation of these methods, the probabilistic inference procedure will search for a prior probability distribution (i.e., an influence of the form $A|_p$, no conditioning propositions) before attempting to use probabilistic influences of the form $A|_pB$, which in general can increase the size of the probabilistic representation. Thus, the scheme embodies a search for a minimum size probabilistic model.

The modular nature of the influences and inference procedures also make it possible to obtain multiple probabilistic models of the same phenomena. That is, within a given knowledge base it may be possible to construct several probabilistic models which will provide results for a proposition of interest. These may represent different conceptions of the world, levels of abstraction, or model sizes based on computational considerations. This raises the issue of consistency.

If multiple models or probability distributions are consistent with a single knowledge base and query, how can we resolve these differences? One answer is further refinement of the knowledge base to make explicit any conditions under which one representation is preferred to another. Similarly, one can construct probabilistic or other representations which embody methods for combining information and outputs. The inference methods as currently defined do not explicitly have methods for resolving and insuring consistency among the multiple models which may be derivable from a knowledge base. However, the overall approach provides the ability to **generate** the multiple models and then allow a higher level authority, perhaps the human decision maker, to interpret and integrate the findings. Development of explicit and formal methods for reasoning about the use of alternative models and their conclusions is an area of future research.

---

[2] The construction of any probabilistic model and the assessment of probabilities presumes a background state of information or knowledge. The set of logically derivable conclusions from the knowledge base at the time of a probabilistic inference make up this state of information.

## IV. SUMMARY

An integrated first-order for representing logical, probabilistic, and decision-theoretic constructs, as well as a set of inference techniques which operate over the language, have been developed. Inference is based on the dynamic construction of probabilistic networks for generation of probabilistic and decision-theoretic conclusions. The approach allows reasoning regarding the domain, as well as reasoning about the construction of a probabilistic or decision-theoretic model addressing a particular query.